\documentclass[letterpaper, 10 pt, journal]{ieeeconf}

\IEEEoverridecommandlockouts
\usepackage{cite}

\usepackage{multicol}
\usepackage{amsmath}
\usepackage{amssymb}
\usepackage{graphicx}
\usepackage{dsfont}
\usepackage{multirow}
\usepackage{diagbox}
\usepackage[table]{xcolor}
\usepackage{cleveref}
\usepackage{booktabs}
\usepackage{microtype}
\usepackage{url}

\makeatletter
\let\MYcaption\@makecaption
\makeatother
\usepackage[font=footnotesize]{subcaption}
\makeatletter
\let\@makecaption\MYcaption
\makeatother

\begin{document}

\title{\LARGE \bf Automatic Map Density Selection for Locally-Performant Visual Place Recognition
}

\author{Somayeh Hussaini, Tobias Fischer, and Michael Milford
\thanks{The authors are with the QUT Centre for Robotics, Queensland University of Technology, Brisbane, QLD 4000, Australia (e-mail: {\tt\footnotesize s.hussaini@qut.edu.au}).}%
}

\maketitle
\thispagestyle{empty}
\pagestyle{empty}

\begin{abstract}

A key challenge in translating Visual Place Recognition (VPR) from the lab to long-term deployment is ensuring a priori that a system can meet user-specified performance requirements across different parts of an environment, rather than just on average globally. One critical mechanism for controlling this local performance is the density of the reference mapping database, yet this factor is largely neglected in existing work, where fixed, engineering-driven sampling densities based on sensors, storage, or GPS frequency are typically used. In this paper, we propose a VPR mapping approach that uses two reference traverses from the operating environment to automatically select an appropriate map density satisfying two user-defined requirements: (1) a target Local Recall@1 level, and (2) the proportion of the operational environment over which it must be met or exceeded, which we term the Recall Achievement Rate (RAR). Our approach uses spatial consistency and coherence features from reference-to-reference matches at multiple map densities to estimate the density needed to meet these targets on unseen deployment data. Through experiments across two VPR methods and the Nordland and Oxford RobotCar benchmarks, we show that our system consistently meets or exceeds the target Local Recall@1 over at least the user-specified proportion of the environment. Comparisons with alternative baselines show that it reliably selects an appropriate operating point in map density, avoiding unnecessarily dense maps. Finally, ablations evaluate sensitivity to reference traversal choice and segment length, and our analysis reveals that conventional global Recall@1 is a poor predictor of the often more operationally meaningful RAR metric.

\end{abstract}

\section{Introduction}

Visual Place Recognition (VPR) is the task of determining whether a query place has been visited previously by identifying its match within a database of reference places~\cite{Lowry2015, masone2021survey, zhang2021visual, garg2021your}. 
It is studied both as a standalone problem and as a component of Simultaneous Localisation And Mapping (SLAM) and visual localisation systems, where modern machine learning methods achieve strong benchmark performance~\cite{masone2021survey, zhang2021visual, berton2025megaloc}.

However, strong benchmark performance does not necessarily translate into reliable deployment. The commonly reported global Recall@1, the fraction of queries whose highest-ranked match lies within the ground-truth tolerance of the true place, is an aggregate measure that does not capture performance variation across an environment. Two systems can achieve the same global Recall@1 while one performs consistently across the route and the other performs well only in selected regions and fails catastrophically elsewhere.

For operational settings, it is often more useful to specify a target Local Recall@1 within individual environmental segments. We define the Recall Achievement Rate (RAR) as the proportion of segments in which this target is met or exceeded. As we show, \textbf{RAR is poorly predicted by global Recall@1}. A further challenge is selecting a system configuration that satisfies the desired Local Recall@1 and RAR \emph{before} deployment, rather than identifying it retrospectively.

A key mechanism for controlling local VPR performance is the density of the reference database. Denser maps may improve matching but increase storage, map construction, and retrieval cost, which for exhaustive search scales linearly with the number of reference descriptors. Existing systems typically fix density from engineering considerations such as sensor frequency, storage capacity, or GPS availability, rather than explicit local performance requirements.

\begin{figure}[t]
\centerline{\includegraphics[width=1.0\columnwidth]{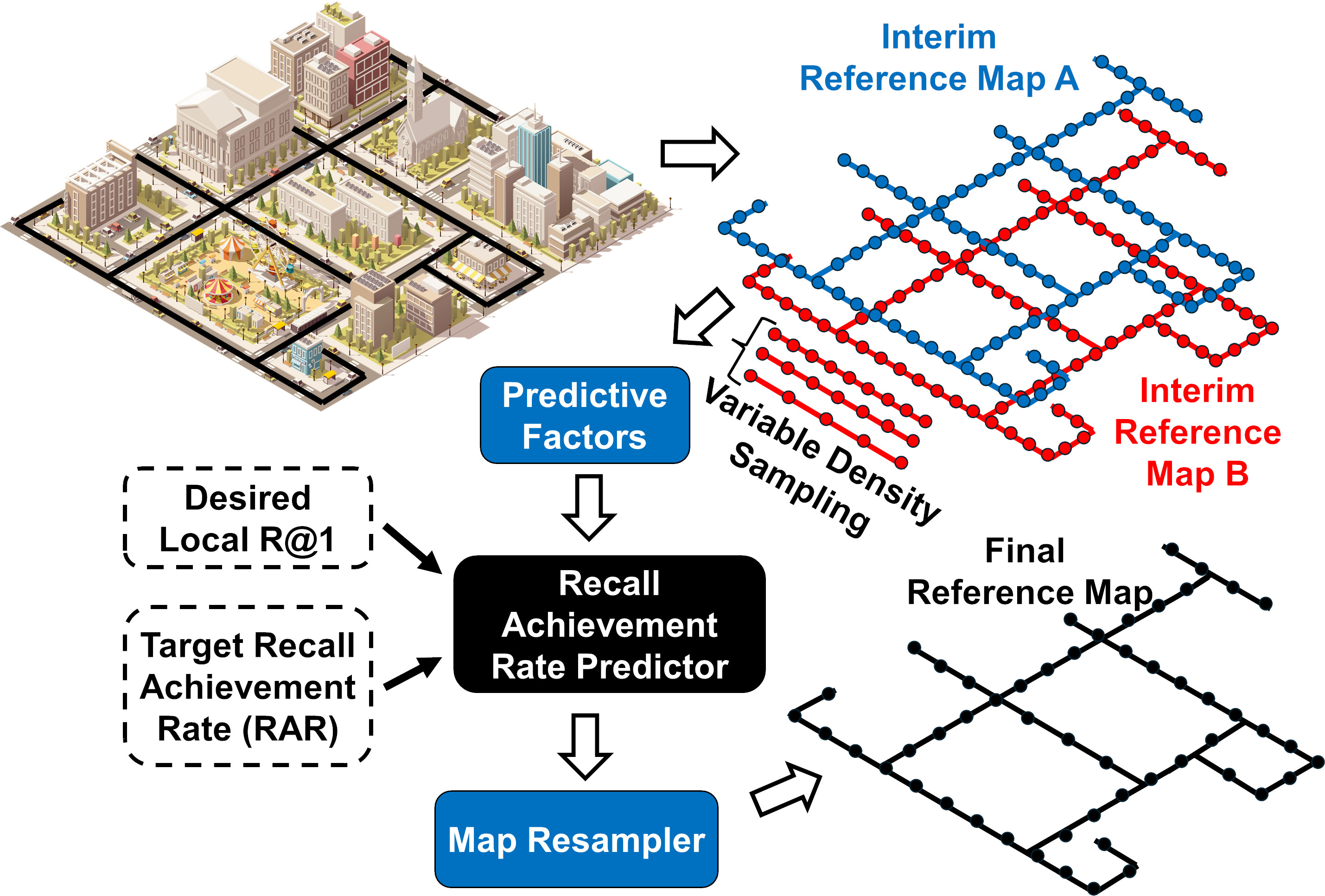}}
\caption{Our approach models the relationship between segment-wise Recall@1 and features extracted from matching two reference traverses at multiple candidate map densities. Given a user-specified Local Recall@1 threshold and target Recall Achievement Rate (RAR), it selects either a uniform or per-segment reference density predicted to satisfy the threshold over the required proportion of environmental segments.
\vspace*{-0.3cm}
}
\label{fig:intro_schematic}
\end{figure}

To address this, we present an automatic VPR map density selection approach using two reference traverses from the operating environment. At each candidate density, it extracts features describing the spatial consistency and coherence of reference-to-reference matches and uses them to estimate segment-level Recall@1. This estimation step is modular: our primary instantiation is a learned predictor, but a simple lookup table is similarly competitive, indicating that the useful signal comes from the reference-to-reference matching rather than the specific mechanism. These estimates select the sparsest reference map expected to satisfy the user-specified Local Recall@1 and RAR requirements on subsequent query data, as illustrated in~\Cref{fig:intro_schematic}. 
The approach targets repeated-route applications such as logistics, inspection, and autonomous vehicles, where multiple traversals accumulate naturally and the second traversal supplies local matching difficulty, replacing manual tuning or conservative oversampling.

We consider two selection policies. The constant selector applies one subsampling factor across the environment, while the variable selector assigns a separate density to each segment. Although the latter adapts directly to local difficulty, both perform comparably at equal reference budgets, reducing their practical distinction to a trade-off between adaptive storage allocation and deployment simplicity.

Across MixVPR and CosPlace on the Nordland and Oxford RobotCar benchmarks, both selectors satisfy a wide range of user-specified RAR targets across different Local Recall@1 thresholds while avoiding unnecessarily dense reference maps, and outperform fixed-density baselines in balancing target satisfaction and reference efficiency.

\noindent The main contributions of this work are:
\begin{enumerate}
    \item We formulate automatic VPR map density selection using user-specified Local Recall@1 and RAR requirements.
    \item We introduce a segment-level recall estimation pipeline, instantiated with both a learned predictor and a simple lookup table, coupled with constant and variable density selection policies.
    \item We evaluate the approach across two VPR methods and two long-term localisation benchmarks, demonstrating reliable target satisfaction and showing that global Recall@1 is a poor proxy for local performance.
\end{enumerate}

\section{Related Work}

Our work draws on two areas: sampling-density selection in SLAM and robot navigation, and reference-database adaptation in Visual Place Recognition (VPR). 
Modern VPR relies on learned global descriptors that retrieve robustly under appearance and viewpoint change, e.g. MixVPR~\cite{ali2023mixvpr} and CosPlace~\cite{berton2022rethinking}, both of which we adopt. 
These advances improve \emph{descriptor quality} and are complementary to our focus.
Given a capable descriptor, we study how reference-database density governs \emph{local} recall. To our knowledge, selecting reference density specifically to meet user-defined local performance requirements has not been systematically studied in VPR.

\subsection{Sampling techniques in SLAM and robot navigation}

In SLAM and visual odometry, keyframe selection trades computational cost against robustness. 
Adaptive strategies use cues such as IMU-based motion~\cite{piao2019real}, variable-density submap construction~\cite{chen2023rumination}, and inter-frame pose analysis with fuzzy inference~\cite{zu2024adaptive}. 
In LiDAR-SLAM, keyframes are chosen from environment spaciousness~\cite{chen2022direct}, information-theoretic criteria~\cite{zeng2023entropy}, or redundancy-minimising combinatorial~\cite{stathoulopoulos2026minimal} and submodular~\cite{thorne2025submodular} optimisation. 
These vary local density from the mapping process itself rather than from explicit performance targets. 
In contrast, we select reference density to satisfy user-specified local requirements, supporting both a constant density and a per-segment variable density, in each case choosing the sparsest sampling estimated to meet the target Local Recall@1 and RAR.

\subsection{Reference database adaptation in VPR}

Several VPR works modify the reference database with objectives different from ours. 
MRS-VPR~\cite{yin2019mrs} refines matches from coarse-to-fine via multi-resolution downsampling and particle filtering.
Bag of Sampled Words~\cite{lee2019bag} instead samples which intra-image features to keep for robustness under appearance change.
Other works prioritise information content over storage. 
Bayesian Selective Fusion~\cite{molloy2020intelligent} selects and fuses informative reference images to improve matching accuracy, 
while~\cite{malone2025hyperdimensional} fuses reference images across sets recorded under different conditions.
These methods mainly improve matching robustness or retrieval speed, with some operating at query time. 
Conversely, we infer an appropriate sampling density from per-segment matching characteristics at map-construction time, to meet user-specified local performance requirements while reducing storage.

\section{Methodology}

We study how reference map density affects Visual Place Recognition (VPR) performance, both locally and globally, where each segment represents a local region spanning a fixed physical distance.
Our goal is to select a reference subsampling factor that meets user-specified local performance requirements while reducing the number of stored reference images.
We propose an automatic, model-agnostic reference density selector that operates at the segment level.

Given a user-specified target Local Recall@1, $R_{\mathrm{target}}$, and target Recall Achievement Rate, $\mathrm{RAR}_{\mathrm{target}}$, it returns the \emph{sparsest} sampling predicted to achieve $R_{\mathrm{target}}$ in at least the required proportion of segments. This is expressed as either a single constant subsampling factor $k^{\ast}$ applied uniformly across the traversal, termed the constant selector, or per-segment factors ${k_i^{\ast}}$ that adapt to local difficulty, termed the variable selector.
Both selectors can be applied across VPR systems without model-specific modification.

\begin{figure*}[htpb]
    \centering
    \vspace*{0.1cm}
    \includegraphics[width=0.8\textwidth, trim=1 10 1 0, clip]{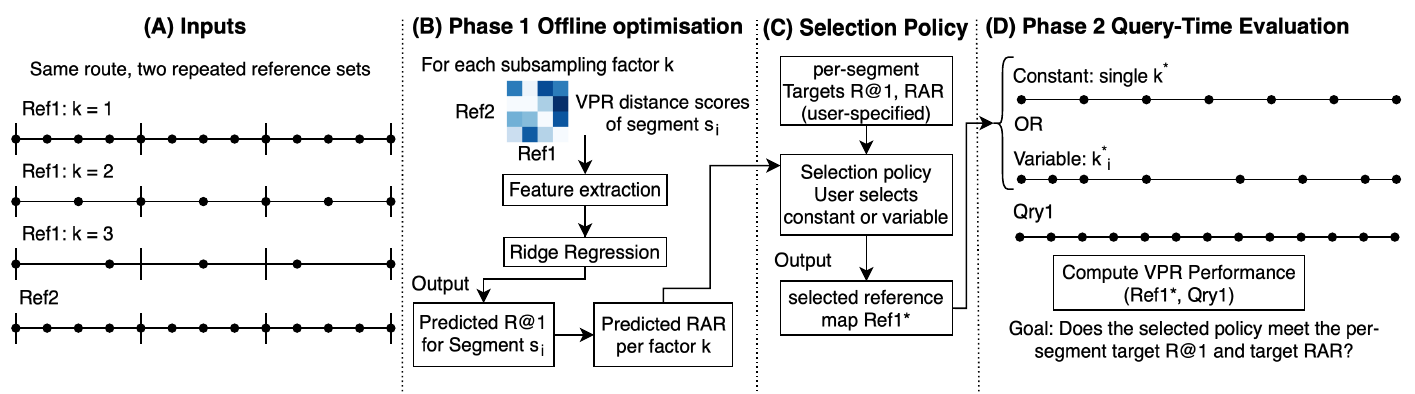}
    \vspace*{-0.3cm}
    \caption{
    Overview of our automatic reference density selection framework.
    (A) Two aligned reference traversals are segmented and processed at multiple candidate densities.
    (B) Features extracted from the Ref1-to-Ref2 distance matrix are used to train density-specific Ridge regressors that predict segment-level Recall@1.
    (C) Given user-specified Local Recall@1 and Recall Achievement Rate (RAR) targets, the selected policy determines either a single constant factor $k^{\ast}$ or per-segment factors $\{k_i^{\ast}\}$.
    (D) The resulting reference map Ref1$^{\ast}$ is evaluated against the unseen query traversal Qry1.
    }
    \label{fig:methodology}
    \vspace*{-0.3cm}
\end{figure*}

\subsection{Problem formulation}

We assume that the VPR dataset consists of repeated traversals corresponding to a robot traversing the same route.
Our setup includes two spatially aligned reference traversals, Ref1 and Ref2, with one-to-one correspondence between their frames, and one held-out query traversal, Qry1, aligned to the same route for evaluation.

We split each reference traversal into $N$ segments, $\mathcal{S} = \{s_1, \ldots, s_N\}$, where each segment $s_i$ covers approximately $d$ metres of physical distance. The segments form a fixed, contiguous, non-overlapping partition determined solely by $d$, which is ablated in~\Cref{res:abl_segment_len}.
We denote the set of $M$ reference subsampling factors as:
    $\mathcal{K} = \{k_1, k_2, \ldots, k_M\},$
where $k \in \mathcal{K}$ indicates sampling every $k$-th reference frame, such that $k=1$ uses all frames and $k=2$ uses every second frame and so forth. Query sampling is fixed at $k=1$ to ensure comprehensive coverage.

\subsubsection{Recall Achievement Rate (RAR)}

Given per-segment Recall@1 values ${R_{i,k}}$ at subsampling factor $k$ and the user-specified target $R_{\mathrm{target}}$, the Recall Achievement Rate is the fraction of segments that meet or exceed the target:
\vspace*{-0.2cm}
\begin{equation}
    \mathrm{RAR}_k =
    \frac{1}{N}
    \sum_{i=1}^{N}
    \mathds{1}[R_{i,k} \geq R_{\mathrm{target}}],
    \label{eq:rar}
    \vspace*{-0.2cm}
\end{equation}
where $\mathds{1}[\cdot]$ is the indicator function.
RAR captures the proportion of the environment that meets the required local recall, complementing global Recall@1, which captures only average performance.
We instantiate Eq.~\eqref{eq:rar} using predicted recalls to obtain an \emph{estimated} RAR for selection and ground-truth recalls on Qry1 to obtain the \emph{achieved} RAR for evaluation. 

\subsection{Reference density selection pipeline}
\label{method:process}

Our pipeline, shown in~\Cref{fig:methodology}, has an offline selection phase using Ref1 and Ref2 and a query-time evaluation phase using Ref1$^{\ast}$ (the subsampled version of Ref1) and Qry1.
Offline, for each subsampling factor $k \in \mathcal{K}$, we compute Ref1-to-Ref2 distance scores using the base VPR model. For every segment $s_i$, we extract a feature vector $\mathbf{x}_{i,k}$ from the distance matrix (\Cref{method:feat}) together with the segment's ground-truth Recall@1, and train a Ridge-regression predictor $f_k$ (\Cref{method:pred}) whose output $\hat{R}_{i,k}$ predicts the segment's Recall@1.
The selector then applies either the constant or variable policy (\Cref{method:constant,method:variable}), and the resulting constant factor $k^{\ast}$ or per-segment factors ${k_i^{\ast}}$ are applied to Ref1 to construct the reduced database Ref1$^{\ast}$, against which Qry1 is matched to obtain the achieved Recall@1 and RAR.

Although $f_k$ is fitted and applied to the same Ref1-to-Ref2 segment pairs, Ridge regularisation and the small feature count ($F=4$) limit model complexity. The fitted predictions are used only for density selection and are never reported as system performance. All reported performance is measured using the independent Ref1$^{\ast}$ and Qry1 pair, with Qry1 unseen during selection.

\subsection{Feature extraction for the Recall@1 predictor}
\label{method:feat}

The base VPR model encodes each frame into a global descriptor. The VPR distance matrix $\mathbf{D}$ holds the pairwise descriptor distances between Ref1 and Ref2 frames, from which top-1 retrievals are obtained. 
For each segment $s_i$ at subsampling factor $k$, we extract $F=4$ features from the VPR distance matrix, capturing the spatial structure and quality of place matches. 
Ref2 acts as the probe against the Ref1 database. Within each segment, we retrieve the top-1 Ref1 match for each Ref2 frame $j = \{ 1, \ldots, n_{q,i} \}$, where $n_{q,i}$ is the number of Ref2 frames in segment $s_i$.
Each retrieval has a route position $p_{i,j}$ in metres from the densest reference sampling. We define the feature vector $\mathbf{x}_{i,k} = [x_{i,k,1}, x_{i,k,2}, x_{i,k,3}, x_{i,k,4}]^T$ as follows.

\subsubsection{Jump Rate ($x_{i,k,1}$)}

Fraction of consecutive predictions with large spatial discontinuities:
\vspace*{-0.2cm}
\begin{equation}
    x_{i,k,1} =
    \frac{1}{n_{q,i} - 1}
    \sum_{j=1}^{n_{q,i}-1}
    \mathds{1}[|p_{i,j+1} - p_{i,j}| > \tau],
    \vspace*{-0.2cm}
\end{equation}
where $\tau$ is the ground-truth tolerance in metres. High Jump Rates suggest inconsistent place matching.

\subsubsection{Fraction Outside Main Cluster ($x_{i,k,2}$)}

Proportion of predictions outside the dominant spatial region. We discretise positions into bins of width $d$ (the segment distance), take the most frequent bin $b_i^{\ast}$, and compute:

\vspace*{-0.2cm}
\begin{equation}
    x_{i,k,2} =
    1 -
    \frac{1}{n_{q,i}}
    \sum_{j=1}^{n_{q,i}}
    \mathds{1}
    [|b_{i,j} - b_i^{\ast}| \leq 1],
    \vspace*{-0.2cm}
\end{equation}
where $b_{i,j} = \lfloor p_{i,j}/d \rfloor$. Higher values indicate spatially scattered predictions.

\subsubsection{Largest Cluster Fraction ($x_{i,k,3}$)}

Proportion of predictions in the largest spatially coherent cluster:
\vspace*{-0.1cm}
\begin{equation}
    x_{i,k,3} =
    \frac{\max_c |C_{i,c}|}{n_{q,i}},
    \vspace*{-0.2cm}
\end{equation}
where $C_{i,c}$ is a cluster of predictions with consecutive positions within tolerance $\tau$, formed by sorting positions and splitting where consecutive positions differ by more than $\tau$. High values indicate spatially concentrated, consistent place matching.

\subsubsection{Turn Rate ($x_{i,k,4}$)}

Non-linear behaviour in the prediction sequence, measured using the second-order difference:
\vspace*{-0.2cm}
\begin{equation}
    x_{i,k,4} =
    \frac{1}{n_{q,i} - 2}
    \sum_{j=1}^{n_{q,i}-2}
    \mathds{1}
    \left[
    p_{i,j+2} - 2p_{i,j+1} + p_{i,j} \neq 0
    \right].
    \label{eq:turnrate}
\end{equation}
Since $p_{i,j}$ is discrete, the second-order difference is zero for a locally linear progression and non-zero otherwise, so a high Turn Rate flags frequent changes in matching progression and potential instability.

\subsection{Segment-level Recall@1 predictor}
\label{method:pred}

We use Ridge regression, an L2-regularised linear model, to predict per-segment Recall@1. It is well suited to the small, potentially correlated feature set.
For each subsampling factor $k$, the Ridge-regression model is:
\begin{equation}
    f_k(\mathbf{x}_{i,k}) =
    \mathbf{w}_k^T \mathbf{x}_{i,k} + b_k,
    \vspace*{-0.1cm}
\end{equation}
with weights $\mathbf{w}_k \in \mathbb{R}^F$ and bias $b_k \in \mathbb{R}$, learned by minimising:
\vspace*{-0.1cm}
\begin{equation}
    \mathbf{w}_k^*, b_k^* = \arg\min_{\mathbf{w}_k, b_k} \sum_{i=1}^{N} (R_{i,k} - \mathbf{w}_k^T \mathbf{x}_{i,k} - b_k)^2 + \lambda \|\mathbf{w}_k\|_2^2,
    \vspace*{-0.2cm}
\end{equation}
where $R_{i,k}$ is the ground-truth Recall@1 for segment $s_i$ at factor $k$ on Ref1 and Ref2, and $\lambda$ controls the trade-off between fit and model complexity.
The output $\hat{R}_{i,k} = f_k(\mathbf{x}_{i,k})$ is the predicted Recall@1 for segment $s_i$ at factor $k$. This predictor stage is one instantiation of a general recall-estimation step. A direct Ref1-to-Ref2 lookup is an alternative, evaluated in~\Cref{res:pred_and_lookup}.

\subsection{Constant density selection policy}
\label{method:constant}

The predicted recalls $\hat{R}_{i,k}$ define, for each factor, the \emph{estimated} Recall Achievement Rate:
\vspace*{-0.1cm}
\begin{equation}
    \widehat{\mathrm{RAR}}_k =
    \mathrm{RAR}({\hat{R}_{i,k}}).
    \label{eq:rarhat}
    \vspace*{-0.2cm}
\end{equation}
The constant selector finds all feasible factors whose estimated RAR meets the target, then selects the sparsest, corresponding to the largest $k$. If none are feasible, it selects the factor with the highest estimated RAR:
\vspace*{-0.1cm}
\begin{equation}
k^{\ast} =
\begin{cases}
    \max \{\, k \in \mathcal{K} :
    \widehat{\mathrm{RAR}}_k
    \geq
    \mathrm{RAR}_{\mathrm{target}} \,\},
    & \text{if feasible},
    \\
    \arg\max_{k \in \mathcal{K}}
    \widehat{\mathrm{RAR}}_k,
    & \text{otherwise}.
\end{cases}
\end{equation}
Here, feasible means that at least one $k \in \mathcal{K}$ satisfies
$\widehat{\mathrm{RAR}}_k \geq \mathrm{RAR}_{\mathrm{target}}$. Ties in the fallback case are resolved in favour of the largest $k$. 
The selected $k^{\ast}$ is applied uniformly across all segments to construct Ref1$^{\ast}$.

\subsection{Variable density selection policy}
\label{method:variable}

The variable selector assigns each segment its own subsampling factor in two stages. First, every segment receives a baseline density. Second, segments with the most reliable predictions are promoted towards a margin-inflated target.

Let $\hat{R}_i^{\max} = \max_{k \in \mathcal{K}} \hat{R}_{i,k}$ denote the maximum predicted Recall@1 for segment $s_i$ across all candidate subsampling factors. 
Each segment first takes the sparsest factor whose predicted Recall@1 meets the adaptive floor
\mbox{$R^{\mathrm{floor}} = R_{\mathrm{target}} \cdot \mathrm{RAR}_{\mathrm{target}}$}.
Segments selected for promotion instead aim for the effective target:
\vspace*{-0.1cm}
\begin{equation}
    R_i^{\mathrm{eff}} =
    \min(\hat{R}_i^{\max}, R_{\mathrm{target}} + \epsilon_R),
    \vspace*{-0.1cm}
\end{equation}
where $\epsilon_R \geq 0$ absorbs Ref2-to-Qry1 performance variation, while the ceiling prevents targeting recall that a segment is not predicted to reach.
The inflated RAR target,
$\mathrm{RAR}^{\mathrm{infl}} =
\min(1, \mathrm{RAR}_{\mathrm{target}} + \epsilon_{\mathrm{RAR}})$,
determines how many segments are promoted.
This count is capped by the number predicted to reach the original recall target:
\vspace*{-0.1cm}
\begin{equation}
    n_\text{feas} = | \{ i : \hat{R}_i^{\max} \geq R_\text{target} \} |,
    n_\text{req} = \min ( n_\text{feas},\, \lceil \text{RAR}^\text{infl}\, N \rceil ).
    \vspace*{-0.05cm}
\end{equation}
We promote the $n_{\mathrm{req}}$ segments with the highest reliability
$\rho_i = \mathrm{margin}_i \cdot \mathrm{breadth}_i$,
where $\mathrm{margin}_i$ is the predicted headroom above $R_i^{\mathrm{eff}}$ and $\mathrm{breadth}_i$ is the fraction of factors in $\mathcal{K}$ that meet it.
Each promoted segment takes the sparsest factor meeting $R_i^{\mathrm{eff}}$; the remaining segments retain their baseline density. Coverage is always evaluated against the original $R_{\mathrm{target}}$.

\section{Experimental Setup}
 
\subsection{Datasets}
 
\subsubsection{Nordland dataset}
The Nordland dataset~\cite{sunderhauf2013we} captures a 728 km train journey in Norway recorded across four seasons.
Following standard practice~\cite{hausler2021patch, camara2020visual}, we remove sections where train speed falls below 15 km/h using the provided GPS data. We use the autumn and spring traverses as Ref1 and Ref2, and the summer traverse as Qry1. We retain the first 20,000 spatially aligned images with segment distance $d = 7,900$ m, corresponding to approximately 200 frames per segment and yielding 100 segments. Consecutive images are approximately 39 m apart, and we use a ground-truth tolerance of $\pm 1,990$ m, corresponding to approximately 50 frames.
 
\subsubsection{Oxford RobotCar dataset}
The Oxford RobotCar dataset~\cite{maddern20171} contains over 100 traverses of Oxford city captured under varying conditions, including different times of day and seasons. Following~\cite{molloy2020intelligent, hussaini2022spiking}, we use front-left stereo frames from the Sun (2015-08-12-15-04-18) and Rain (2015-10-29-12-18-17) traverses as Ref1 and Ref2, and the Dusk (2014-11-21-16-07-03) traverse as Qry1. The traverses are spatially aligned and sampled at approximately one image per metre, yielding 3,800 matched places per traversal. With segment distance $d = 50$ m, the retained route is divided into 76 segments. We use a ground-truth tolerance of $\pm 25$ m.

\subsection{Implementation and evaluation protocol}
\label{exp_setup:configs}
 
During evaluation, each Qry1 frame is matched against the reduced database Ref1$^{\ast}$, constructed using either the selected constant factor $k^{\ast}$ or the per-segment factors ${k_i^{\ast}}$. This measures how well the density policy selected from Ref1 and Ref2 transfers to the unseen Qry1 traversal.
We evaluate two VPR methods, MixVPR~\cite{ali2023mixvpr} and CosPlace~\cite{berton2022rethinking}, using the implementations in~\cite{berton2023eigenplaces}.
For the Ridge-regression predictor, we set the regularisation parameter $\lambda$ to 0.01, with an ablation provided in~\Cref{res:abl_pred_hyp}.

For the variable selector, we use margins $\epsilon_R = 0.07$ and $\epsilon_{\mathrm{RAR}} = 0.10$. These are calibrated once using Nordland's winter traversal as held-out validation data to estimate the Ref2-to-Qry1 performance shift; the winter traversal is excluded from all reported evaluation. The same values are then used across both VPR methods and datasets.

To isolate the effect of density allocation from total storage, we also evaluate a budget-equalised variable selector whose final reference count is matched to that of the constant selector. If the variable selector exceeds this budget, segment densities are reduced in order of the largest reference-count savings; if it falls below the budget, segment densities are increased in order of the smallest reference-count additions until the total reference counts match.

For Ref1, we evaluate the candidate factors
$\mathcal{K} = \{1,2,3,4,5,10,15,20,25,30,35,40,45,50\}$,
while Ref2 and Qry1 remain at full density, $k=1$. We compare against all fixed global densities $k \in \mathcal{K}$, where one factor is applied uniformly across the full traversal, as shown in~\Cref{fig:summary_fig_v2}. In~\Cref{tab:rar_absolute_with_refs}, we use $k=4$ as a representative fixed-density baseline for detailed comparison.

\subsection{Recall Achievement Rate evaluation}

All reported RAR values are computed via Eq.~\eqref{eq:rar} using ground-truth Recall@1 on the unseen Ref1$^{\ast}$ and Qry1 pair, evaluated at the selected constant factor $k^{\ast}$ or per-segment variable factors $k_i^{\ast}$.
These achieved values mirror the estimated RAR used during selection, but are computed only from unseen Qry1 data.

\section{Results}

\begin{figure}[t]
\centering
\vspace*{0.2cm}
{\includegraphics[width=0.8\linewidth]{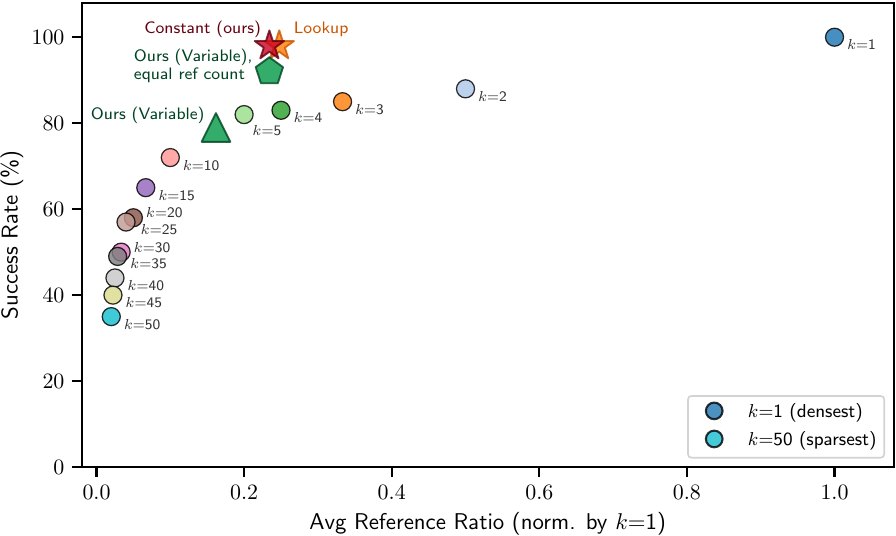}}
\vspace*{-0.2cm}
\caption{
Success rate versus average reference ratio across all target Local Recall@1 and Recall Achievement Rate (RAR) combinations, VPR methods, and datasets. Success rate is the proportion of configurations for which the achieved RAR meets or exceeds its target. The reference ratio is the retained reference count normalised by the densest map at $k=1$. Circles denote fixed global densities, while the star, triangle, and pentagon denote the constant, variable, and budget-equalised variable selectors, respectively.
\vspace*{-0.4cm}}
\label{fig:summary_fig_v2}
\end{figure}

\begin{table*}[htbp]
\centering
\vspace*{0.4cm}
\caption{Achieved RAR values. \textbf{Top block}: achieved RAR per cell; \colorbox{blue!20}{blue} = above target RAR threshold, \colorbox{red!25}{red} = below. {\boldmath$*$} indicates the cell is at its maximum achievable RAR (ceiling) despite falling below the target threshold. \textbf{Bottom block}: reference count ratio normalised by $k$=1; \colorbox[rgb]{0.45,0.77,0.46}{\textcolor{black}{green}} = sparsest (fewest refs), \colorbox[rgb]{1.00,1.00,0.90}{pale yellow} = equal to densest ($k$=1). Last column: mean ref ratio (normalised by $k$=1) across RAR thresholds.}
\label{tab:rar_absolute_with_refs}
\setlength{\tabcolsep}{3pt}
\vspace*{-0.25cm}

\begin{minipage}{0.49\textwidth}
\centering
\textbf{(a) Nordland - MixVPR} \\

    \resizebox{\linewidth}{!}{
    \begin{tabular}{c|ccccc|ccccc|cc}
    \hline
     & \multicolumn{5}{c|}{Constant (ours)} & \multicolumn{5}{c|}{globally fixed ($k$ = 4)} & \multicolumn{2}{c}{Success Rate} \\
    \hline
    \diagbox{Target R@1}{Target RAR} & 0.2 & 0.4 & 0.6 & 0.8 & 1.0 & 0.2 & 0.4 & 0.6 & 0.8 & 1.0 & Constant (ours) & Base \\
    \hline
    0.2 & \cellcolor{blue!43}0.63 & \cellcolor{blue!23}0.63 & \cellcolor{blue!13}0.73 & \cellcolor{blue!9}0.90 & 1.00 & \cellcolor{blue!60}1.00 & \cellcolor{blue!60}1.00 & \cellcolor{blue!40}1.00 & \cellcolor{blue!19}1.00 & 1.00 & \textbf{5/5} & \textbf{5/5} \\
0.4 & \cellcolor{blue!12}0.32 & \cellcolor{blue!6}0.47 & \cellcolor{blue!21}0.82 & \cellcolor{blue!16}0.97 & 1.00 & \cellcolor{blue!60}1.00 & \cellcolor{blue!60}1.00 & \cellcolor{blue!40}1.00 & \cellcolor{blue!19}1.00 & 1.00 & \textbf{5/5} & \textbf{5/5} \\
0.6 & \cellcolor{blue!8}0.28 & \cellcolor{blue!25}0.65 & \cellcolor{blue!37}0.97 & \cellcolor{blue!16}0.97 & 1.00 & \cellcolor{blue!60}0.98 & \cellcolor{blue!57}0.98 & \cellcolor{blue!38}0.98 & \cellcolor{blue!17}0.98 & \cellcolor{red!25}0.98 & \textbf{5/5} & 4/5 \\
0.8 & \cellcolor{blue!48}0.68 & \cellcolor{blue!28}0.68 & \cellcolor{blue!8}0.68 & \cellcolor{blue!7}0.88 & 0.98* & \cellcolor{blue!60}0.81 & \cellcolor{blue!41}0.81 & \cellcolor{blue!21}0.81 & 0.81 & \cellcolor{red!25}0.81 & \textbf{5/5} & 4/5 \\
1.0 & \cellcolor{blue!43}0.63 & \cellcolor{blue!23}0.63 & 0.63 & 0.63* & 0.63* & \cellcolor{red!25}0.00 & \cellcolor{red!25}0.00 & \cellcolor{red!25}0.00 & \cellcolor{red!25}0.00 & \cellcolor{red!25}0.00 & \textbf{5/5} & 0/5 \\
\hline
    \multicolumn{1}{c|}{} & \multicolumn{5}{c|}{\textit{\# Refs (Constant (ours))}} & \multicolumn{5}{c|}{\textit{\# Refs (globally fixed ($k$ = 4))}} & \multicolumn{2}{c}{Avg Ratio} \\
    \hline
    0.2 & \cellcolor[rgb]{0.197,0.596,0.315}0.02 & \cellcolor[rgb]{0.197,0.596,0.315}0.02 & \cellcolor[rgb]{0.197,0.596,0.315}0.02 & \cellcolor[rgb]{0.205,0.605,0.321}0.03 & \cellcolor[rgb]{0.249,0.663,0.360}0.10 & \cellcolor[rgb]{0.427,0.755,0.452}0.25 & \cellcolor[rgb]{0.427,0.755,0.452}0.25 & \cellcolor[rgb]{0.427,0.755,0.452}0.25 & \cellcolor[rgb]{0.427,0.755,0.452}0.25 & \cellcolor[rgb]{0.427,0.755,0.452}0.25 & 0.04 & 0.25 \\
0.4 & \cellcolor[rgb]{0.205,0.605,0.321}0.03 & \cellcolor[rgb]{0.208,0.610,0.324}0.04 & \cellcolor[rgb]{0.227,0.634,0.340}0.07 & \cellcolor[rgb]{0.249,0.663,0.360}0.10 & \cellcolor[rgb]{1.000,1.000,0.898}1.00 & \cellcolor[rgb]{0.427,0.755,0.452}0.25 & \cellcolor[rgb]{0.427,0.755,0.452}0.25 & \cellcolor[rgb]{0.427,0.755,0.452}0.25 & \cellcolor[rgb]{0.427,0.755,0.452}0.25 & \cellcolor[rgb]{0.427,0.755,0.452}0.25 & 0.25 & 0.25 \\
0.6 & \cellcolor[rgb]{0.227,0.634,0.340}0.07 & \cellcolor[rgb]{0.249,0.663,0.360}0.10 & \cellcolor[rgb]{0.366,0.725,0.421}0.20 & \cellcolor[rgb]{0.366,0.725,0.421}0.20 & \cellcolor[rgb]{1.000,1.000,0.898}1.00 & \cellcolor[rgb]{0.427,0.755,0.452}0.25 & \cellcolor[rgb]{0.427,0.755,0.452}0.25 & \cellcolor[rgb]{0.427,0.755,0.452}0.25 & \cellcolor[rgb]{0.427,0.755,0.452}0.25 & \cellcolor[rgb]{0.427,0.755,0.452}0.25 & 0.31 & 0.25 \\
0.8 & \cellcolor[rgb]{0.366,0.725,0.421}0.20 & \cellcolor[rgb]{0.366,0.725,0.421}0.20 & \cellcolor[rgb]{0.366,0.725,0.421}0.20 & \cellcolor[rgb]{0.526,0.801,0.496}0.33 & \cellcolor[rgb]{1.000,1.000,0.898}1.00 & \cellcolor[rgb]{0.427,0.755,0.452}0.25 & \cellcolor[rgb]{0.427,0.755,0.452}0.25 & \cellcolor[rgb]{0.427,0.755,0.452}0.25 & \cellcolor[rgb]{0.427,0.755,0.452}0.25 & \cellcolor[rgb]{0.427,0.755,0.452}0.25 & 0.39 & 0.25 \\
1.0 & \cellcolor[rgb]{1.000,1.000,0.898}1.00 & \cellcolor[rgb]{1.000,1.000,0.898}1.00 & \cellcolor[rgb]{1.000,1.000,0.898}1.00 & \cellcolor[rgb]{1.000,1.000,0.898}1.00 & \cellcolor[rgb]{1.000,1.000,0.898}1.00 & \cellcolor[rgb]{0.427,0.755,0.452}0.25 & \cellcolor[rgb]{0.427,0.755,0.452}0.25 & \cellcolor[rgb]{0.427,0.755,0.452}0.25 & \cellcolor[rgb]{0.427,0.755,0.452}0.25 & \cellcolor[rgb]{0.427,0.755,0.452}0.25 & 1.00 & 0.25 \\
\hline
    \end{tabular}}
    
\end{minipage}
\hfill
\begin{minipage}{0.49\textwidth}
\centering
\textbf{(b) Nordland - CosPlace} \\

    \resizebox{\linewidth}{!}{
    \begin{tabular}{c|ccccc|ccccc|cc}
    \hline
     & \multicolumn{5}{c|}{Constant (ours)} & \multicolumn{5}{c|}{globally fixed ($k$ = 4)} & \multicolumn{2}{c}{Success Rate} \\
    \hline
    \diagbox{Target R@1}{Target RAR} & 0.2 & 0.4 & 0.6 & 0.8 & 1.0 & 0.2 & 0.4 & 0.6 & 0.8 & 1.0 & Constant (ours) & Base \\
    \hline
    0.2 & \cellcolor{blue!36}0.57 & \cellcolor{blue!16}0.57 & \cellcolor{blue!29}0.89 & \cellcolor{blue!16}0.97 & 1.00 & \cellcolor{blue!60}1.00 & \cellcolor{blue!60}1.00 & \cellcolor{blue!40}1.00 & \cellcolor{blue!19}1.00 & 1.00 & \textbf{5/5} & \textbf{5/5} \\
0.4 & \cellcolor{blue!18}0.38 & \cellcolor{blue!34}0.74 & \cellcolor{blue!33}0.93 & \cellcolor{blue!18}0.99 & 1.00 & \cellcolor{blue!60}0.99 & \cellcolor{blue!59}0.99 & \cellcolor{blue!39}0.99 & \cellcolor{blue!18}0.99 & \cellcolor{red!25}0.99 & \textbf{5/5} & 4/5 \\
0.6 & \cellcolor{blue!35}0.55 & \cellcolor{blue!49}0.89 & \cellcolor{blue!29}0.89 & \cellcolor{blue!17}0.98 & 0.99* & \cellcolor{blue!60}0.94 & \cellcolor{blue!53}0.94 & \cellcolor{blue!34}0.94 & \cellcolor{blue!13}0.94 & \cellcolor{red!25}0.94 & \textbf{5/5} & 4/5 \\
0.8 & \cellcolor{blue!40}0.60 & \cellcolor{blue!29}0.70 & \cellcolor{blue!28}0.88 & \cellcolor{blue!15}0.96 & 0.96* & \cellcolor{blue!49}0.70 & \cellcolor{blue!29}0.70 & \cellcolor{blue!9}0.70 & \cellcolor{red!25}0.70 & \cellcolor{red!25}0.70 & \textbf{5/5} & 3/5 \\
1.0 & \cellcolor{blue!36}0.56 & \cellcolor{blue!16}0.56 & 0.56* & 0.56* & 0.56* & \cellcolor{red!25}0.00 & \cellcolor{red!25}0.00 & \cellcolor{red!25}0.00 & \cellcolor{red!25}0.00 & \cellcolor{red!25}0.00 & \textbf{5/5} & 0/5 \\
\hline
    \multicolumn{1}{c|}{} & \multicolumn{5}{c|}{\textit{\# Refs (Constant (ours))}} & \multicolumn{5}{c|}{\textit{\# Refs (globally fixed ($k$ = 4))}} & \multicolumn{2}{c}{Avg Ratio} \\
    \hline
    0.2 & \cellcolor[rgb]{0.197,0.596,0.315}0.02 & \cellcolor[rgb]{0.197,0.596,0.315}0.02 & \cellcolor[rgb]{0.205,0.605,0.321}0.03 & \cellcolor[rgb]{0.208,0.610,0.324}0.04 & \cellcolor[rgb]{0.249,0.663,0.360}0.10 & \cellcolor[rgb]{0.427,0.755,0.452}0.25 & \cellcolor[rgb]{0.427,0.755,0.452}0.25 & \cellcolor[rgb]{0.427,0.755,0.452}0.25 & \cellcolor[rgb]{0.427,0.755,0.452}0.25 & \cellcolor[rgb]{0.427,0.755,0.452}0.25 & 0.04 & 0.25 \\
0.4 & \cellcolor[rgb]{0.208,0.610,0.324}0.04 & \cellcolor[rgb]{0.227,0.634,0.340}0.07 & \cellcolor[rgb]{0.249,0.663,0.360}0.10 & \cellcolor[rgb]{0.366,0.725,0.421}0.20 & \cellcolor[rgb]{0.714,0.882,0.574}0.50 & \cellcolor[rgb]{0.427,0.755,0.452}0.25 & \cellcolor[rgb]{0.427,0.755,0.452}0.25 & \cellcolor[rgb]{0.427,0.755,0.452}0.25 & \cellcolor[rgb]{0.427,0.755,0.452}0.25 & \cellcolor[rgb]{0.427,0.755,0.452}0.25 & 0.18 & 0.25 \\
0.6 & \cellcolor[rgb]{0.249,0.663,0.360}0.10 & \cellcolor[rgb]{0.366,0.725,0.421}0.20 & \cellcolor[rgb]{0.366,0.725,0.421}0.20 & \cellcolor[rgb]{0.526,0.801,0.496}0.33 & \cellcolor[rgb]{1.000,1.000,0.898}1.00 & \cellcolor[rgb]{0.427,0.755,0.452}0.25 & \cellcolor[rgb]{0.427,0.755,0.452}0.25 & \cellcolor[rgb]{0.427,0.755,0.452}0.25 & \cellcolor[rgb]{0.427,0.755,0.452}0.25 & \cellcolor[rgb]{0.427,0.755,0.452}0.25 & 0.37 & 0.25 \\
0.8 & \cellcolor[rgb]{0.366,0.725,0.421}0.20 & \cellcolor[rgb]{0.427,0.755,0.452}0.25 & \cellcolor[rgb]{0.714,0.882,0.574}0.50 & \cellcolor[rgb]{1.000,1.000,0.898}1.00 & \cellcolor[rgb]{1.000,1.000,0.898}1.00 & \cellcolor[rgb]{0.427,0.755,0.452}0.25 & \cellcolor[rgb]{0.427,0.755,0.452}0.25 & \cellcolor[rgb]{0.427,0.755,0.452}0.25 & \cellcolor[rgb]{0.427,0.755,0.452}0.25 & \cellcolor[rgb]{0.427,0.755,0.452}0.25 & 0.59 & 0.25 \\
1.0 & \cellcolor[rgb]{1.000,1.000,0.898}1.00 & \cellcolor[rgb]{1.000,1.000,0.898}1.00 & \cellcolor[rgb]{1.000,1.000,0.898}1.00 & \cellcolor[rgb]{1.000,1.000,0.898}1.00 & \cellcolor[rgb]{1.000,1.000,0.898}1.00 & \cellcolor[rgb]{0.427,0.755,0.452}0.25 & \cellcolor[rgb]{0.427,0.755,0.452}0.25 & \cellcolor[rgb]{0.427,0.755,0.452}0.25 & \cellcolor[rgb]{0.427,0.755,0.452}0.25 & \cellcolor[rgb]{0.427,0.755,0.452}0.25 & 1.00 & 0.25 \\
\hline
    \end{tabular}}
    
\end{minipage}

\vspace{3mm}

\begin{minipage}{0.49\textwidth}
\centering
\textbf{(c) Oxford RobotCar - MixVPR} \\

    \resizebox{\linewidth}{!}{
    \begin{tabular}{c|ccccc|ccccc|cc}
    \hline
     & \multicolumn{5}{c|}{Constant (ours)} & \multicolumn{5}{c|}{globally fixed ($k$ = 4)} & \multicolumn{2}{c}{Success Rate} \\
    \hline
    \diagbox{Target R@1}{Target RAR} & 0.2 & 0.4 & 0.6 & 0.8 & 1.0 & 0.2 & 0.4 & 0.6 & 0.8 & 1.0 & Constant (ours) & Base \\
    \hline
    0.2 & \cellcolor{blue!60}1.00 & \cellcolor{blue!60}1.00 & \cellcolor{blue!40}1.00 & \cellcolor{blue!19}1.00 & 1.00 & \cellcolor{blue!60}1.00 & \cellcolor{blue!60}1.00 & \cellcolor{blue!40}1.00 & \cellcolor{blue!19}1.00 & 1.00 & \textbf{5/5} & \textbf{5/5} \\
0.4 & \cellcolor{blue!60}1.00 & \cellcolor{blue!60}1.00 & \cellcolor{blue!40}1.00 & \cellcolor{blue!19}1.00 & 1.00 & \cellcolor{blue!60}1.00 & \cellcolor{blue!60}1.00 & \cellcolor{blue!40}1.00 & \cellcolor{blue!19}1.00 & 1.00 & \textbf{5/5} & \textbf{5/5} \\
0.6 & \cellcolor{blue!60}0.92 & \cellcolor{blue!52}0.92 & \cellcolor{blue!32}0.92 & \cellcolor{blue!18}0.99 & 1.00 & \cellcolor{blue!60}1.00 & \cellcolor{blue!60}1.00 & \cellcolor{blue!40}1.00 & \cellcolor{blue!19}1.00 & 1.00 & \textbf{5/5} & \textbf{5/5} \\
0.8 & \cellcolor{blue!39}0.59 & \cellcolor{blue!19}0.59 & \cellcolor{blue!18}0.79 & \cellcolor{blue!9}0.89 & 1.00 & \cellcolor{blue!60}1.00 & \cellcolor{blue!60}1.00 & \cellcolor{blue!40}1.00 & \cellcolor{blue!19}1.00 & 1.00 & \textbf{5/5} & \textbf{5/5} \\
1.0 & \cellcolor{blue!60}0.95 & \cellcolor{blue!54}0.95 & \cellcolor{blue!34}0.95 & \cellcolor{blue!14}0.95 & \cellcolor{red!25}0.95 & \cellcolor{blue!60}0.99 & \cellcolor{blue!58}0.99 & \cellcolor{blue!38}0.99 & \cellcolor{blue!18}0.99 & 0.99* & 4/5 & \textbf{5/5} \\
\hline
    \multicolumn{1}{c|}{} & \multicolumn{5}{c|}{\textit{\# Refs (Constant (ours))}} & \multicolumn{5}{c|}{\textit{\# Refs (globally fixed ($k$ = 4))}} & \multicolumn{2}{c}{Avg Ratio} \\
    \hline
    0.2 & \cellcolor[rgb]{0.197,0.596,0.315}0.02 & \cellcolor[rgb]{0.197,0.596,0.315}0.02 & \cellcolor[rgb]{0.197,0.596,0.315}0.02 & \cellcolor[rgb]{0.197,0.596,0.315}0.02 & \cellcolor[rgb]{0.249,0.663,0.360}0.10 & \cellcolor[rgb]{0.427,0.755,0.452}0.25 & \cellcolor[rgb]{0.427,0.755,0.452}0.25 & \cellcolor[rgb]{0.427,0.755,0.452}0.25 & \cellcolor[rgb]{0.427,0.755,0.452}0.25 & \cellcolor[rgb]{0.427,0.755,0.452}0.25 & 0.04 & 0.25 \\
0.4 & \cellcolor[rgb]{0.197,0.596,0.315}0.02 & \cellcolor[rgb]{0.197,0.596,0.315}0.02 & \cellcolor[rgb]{0.197,0.596,0.315}0.02 & \cellcolor[rgb]{0.197,0.596,0.315}0.02 & \cellcolor[rgb]{0.249,0.663,0.360}0.10 & \cellcolor[rgb]{0.427,0.755,0.452}0.25 & \cellcolor[rgb]{0.427,0.755,0.452}0.25 & \cellcolor[rgb]{0.427,0.755,0.452}0.25 & \cellcolor[rgb]{0.427,0.755,0.452}0.25 & \cellcolor[rgb]{0.427,0.755,0.452}0.25 & 0.04 & 0.25 \\
0.6 & \cellcolor[rgb]{0.197,0.596,0.315}0.02 & \cellcolor[rgb]{0.197,0.596,0.315}0.02 & \cellcolor[rgb]{0.197,0.596,0.315}0.02 & \cellcolor[rgb]{0.197,0.596,0.315}0.02 & \cellcolor[rgb]{0.249,0.663,0.360}0.10 & \cellcolor[rgb]{0.427,0.755,0.452}0.25 & \cellcolor[rgb]{0.427,0.755,0.452}0.25 & \cellcolor[rgb]{0.427,0.755,0.452}0.25 & \cellcolor[rgb]{0.427,0.755,0.452}0.25 & \cellcolor[rgb]{0.427,0.755,0.452}0.25 & 0.04 & 0.25 \\
0.8 & \cellcolor[rgb]{0.197,0.596,0.315}0.02 & \cellcolor[rgb]{0.197,0.596,0.315}0.02 & \cellcolor[rgb]{0.197,0.596,0.315}0.02 & \cellcolor[rgb]{0.201,0.600,0.318}0.03 & \cellcolor[rgb]{0.249,0.663,0.360}0.10 & \cellcolor[rgb]{0.427,0.755,0.452}0.25 & \cellcolor[rgb]{0.427,0.755,0.452}0.25 & \cellcolor[rgb]{0.427,0.755,0.452}0.25 & \cellcolor[rgb]{0.427,0.755,0.452}0.25 & \cellcolor[rgb]{0.427,0.755,0.452}0.25 & 0.04 & 0.25 \\
1.0 & \cellcolor[rgb]{0.249,0.663,0.360}0.10 & \cellcolor[rgb]{0.249,0.663,0.360}0.10 & \cellcolor[rgb]{0.249,0.663,0.360}0.10 & \cellcolor[rgb]{0.249,0.663,0.360}0.10 & \cellcolor[rgb]{0.249,0.663,0.360}0.10 & \cellcolor[rgb]{0.427,0.755,0.452}0.25 & \cellcolor[rgb]{0.427,0.755,0.452}0.25 & \cellcolor[rgb]{0.427,0.755,0.452}0.25 & \cellcolor[rgb]{0.427,0.755,0.452}0.25 & \cellcolor[rgb]{0.427,0.755,0.452}0.25 & 0.10 & 0.25 \\
\hline
    \end{tabular}}
    
\end{minipage}
\hfill
\begin{minipage}{0.49\textwidth}
\centering
\textbf{(d) Oxford RobotCar - CosPlace} \\

    \resizebox{\linewidth}{!}{
    \begin{tabular}{c|ccccc|ccccc|cc}
    \hline
     & \multicolumn{5}{c|}{Constant (ours)} & \multicolumn{5}{c|}{globally fixed ($k$ = 4)} & \multicolumn{2}{c}{Success Rate} \\
    \hline
    \diagbox{Target R@1}{Target RAR} & 0.2 & 0.4 & 0.6 & 0.8 & 1.0 & 0.2 & 0.4 & 0.6 & 0.8 & 1.0 & Constant (ours) & Base \\
    \hline
    0.2 & \cellcolor{blue!60}1.00 & \cellcolor{blue!60}1.00 & \cellcolor{blue!40}1.00 & \cellcolor{blue!19}1.00 & 1.00 & \cellcolor{blue!60}1.00 & \cellcolor{blue!60}1.00 & \cellcolor{blue!40}1.00 & \cellcolor{blue!19}1.00 & 1.00 & \textbf{5/5} & \textbf{5/5} \\
0.4 & \cellcolor{blue!60}0.99 & \cellcolor{blue!58}0.99 & \cellcolor{blue!38}0.99 & \cellcolor{blue!18}0.99 & 1.00 & \cellcolor{blue!60}1.00 & \cellcolor{blue!60}1.00 & \cellcolor{blue!40}1.00 & \cellcolor{blue!19}1.00 & 1.00 & \textbf{5/5} & \textbf{5/5} \\
0.6 & \cellcolor{blue!58}0.79 & \cellcolor{blue!38}0.79 & \cellcolor{blue!29}0.89 & \cellcolor{blue!18}0.99 & 1.00 & \cellcolor{blue!60}1.00 & \cellcolor{blue!60}1.00 & \cellcolor{blue!40}1.00 & \cellcolor{blue!19}1.00 & 1.00 & \textbf{5/5} & \textbf{5/5} \\
0.8 & \cellcolor{blue!28}0.49 & \cellcolor{blue!35}0.75 & \cellcolor{blue!15}0.75 & \cellcolor{blue!19}1.00 & 1.00 & \cellcolor{blue!60}1.00 & \cellcolor{blue!60}1.00 & \cellcolor{blue!40}1.00 & \cellcolor{blue!19}1.00 & 1.00 & \textbf{5/5} & \textbf{5/5} \\
1.0 & \cellcolor{blue!60}0.86 & \cellcolor{blue!45}0.86 & \cellcolor{blue!25}0.86 & \cellcolor{blue!5}0.86 & \cellcolor{red!25}0.86 & \cellcolor{blue!60}0.91 & \cellcolor{blue!50}0.91 & \cellcolor{blue!30}0.91 & \cellcolor{blue!10}0.91 & \cellcolor{red!25}0.91 & \textbf{4/5} & \textbf{4/5} \\
\hline
    \multicolumn{1}{c|}{} & \multicolumn{5}{c|}{\textit{\# Refs (Constant (ours))}} & \multicolumn{5}{c|}{\textit{\# Refs (globally fixed ($k$ = 4))}} & \multicolumn{2}{c}{Avg Ratio} \\
    \hline
    0.2 & \cellcolor[rgb]{0.197,0.596,0.315}0.02 & \cellcolor[rgb]{0.197,0.596,0.315}0.02 & \cellcolor[rgb]{0.197,0.596,0.315}0.02 & \cellcolor[rgb]{0.197,0.596,0.315}0.02 & \cellcolor[rgb]{0.249,0.663,0.360}0.10 & \cellcolor[rgb]{0.427,0.755,0.452}0.25 & \cellcolor[rgb]{0.427,0.755,0.452}0.25 & \cellcolor[rgb]{0.427,0.755,0.452}0.25 & \cellcolor[rgb]{0.427,0.755,0.452}0.25 & \cellcolor[rgb]{0.427,0.755,0.452}0.25 & 0.04 & 0.25 \\
0.4 & \cellcolor[rgb]{0.197,0.596,0.315}0.02 & \cellcolor[rgb]{0.197,0.596,0.315}0.02 & \cellcolor[rgb]{0.197,0.596,0.315}0.02 & \cellcolor[rgb]{0.205,0.605,0.321}0.03 & \cellcolor[rgb]{0.249,0.663,0.360}0.10 & \cellcolor[rgb]{0.427,0.755,0.452}0.25 & \cellcolor[rgb]{0.427,0.755,0.452}0.25 & \cellcolor[rgb]{0.427,0.755,0.452}0.25 & \cellcolor[rgb]{0.427,0.755,0.452}0.25 & \cellcolor[rgb]{0.427,0.755,0.452}0.25 & 0.04 & 0.25 \\
0.6 & \cellcolor[rgb]{0.197,0.596,0.315}0.02 & \cellcolor[rgb]{0.197,0.596,0.315}0.02 & \cellcolor[rgb]{0.197,0.596,0.315}0.02 & \cellcolor[rgb]{0.216,0.620,0.331}0.05 & \cellcolor[rgb]{0.249,0.663,0.360}0.10 & \cellcolor[rgb]{0.427,0.755,0.452}0.25 & \cellcolor[rgb]{0.427,0.755,0.452}0.25 & \cellcolor[rgb]{0.427,0.755,0.452}0.25 & \cellcolor[rgb]{0.427,0.755,0.452}0.25 & \cellcolor[rgb]{0.427,0.755,0.452}0.25 & 0.04 & 0.25 \\
0.8 & \cellcolor[rgb]{0.197,0.596,0.315}0.02 & \cellcolor[rgb]{0.201,0.600,0.318}0.03 & \cellcolor[rgb]{0.201,0.600,0.318}0.03 & \cellcolor[rgb]{0.249,0.663,0.360}0.10 & \cellcolor[rgb]{0.249,0.663,0.360}0.10 & \cellcolor[rgb]{0.427,0.755,0.452}0.25 & \cellcolor[rgb]{0.427,0.755,0.452}0.25 & \cellcolor[rgb]{0.427,0.755,0.452}0.25 & \cellcolor[rgb]{0.427,0.755,0.452}0.25 & \cellcolor[rgb]{0.427,0.755,0.452}0.25 & 0.05 & 0.25 \\
1.0 & \cellcolor[rgb]{0.249,0.663,0.360}0.10 & \cellcolor[rgb]{0.249,0.663,0.360}0.10 & \cellcolor[rgb]{0.249,0.663,0.360}0.10 & \cellcolor[rgb]{0.249,0.663,0.360}0.10 & \cellcolor[rgb]{0.249,0.663,0.360}0.10 & \cellcolor[rgb]{0.427,0.755,0.452}0.25 & \cellcolor[rgb]{0.427,0.755,0.452}0.25 & \cellcolor[rgb]{0.427,0.755,0.452}0.25 & \cellcolor[rgb]{0.427,0.755,0.452}0.25 & \cellcolor[rgb]{0.427,0.755,0.452}0.25 & 0.10 & 0.25 \\
\hline
    \end{tabular}}
    
\end{minipage}
\vspace*{-0.3cm}
\end{table*}

\subsection{Performance overview}
\label{res:perf_overview}

\Cref{fig:summary_fig_v2} summarises selector success rate against average reference ratio, normalised by the densest map at $k=1$, across all target Local Recall@1 and target RAR combinations, VPR methods, and datasets. Success rate is the proportion of these configurations for which the achieved RAR meets or exceeds its target. We compare against fixed global reference densities, where one value of $k$ is applied uniformly across all segments.
Our constant selector achieves a 98.0\% success rate while retaining only 23.4\% of the densest reference set on average. The densest fixed baseline, $k=1$, reaches 100\% success but retains the complete reference database, while increasingly sparse fixed densities reduce storage at the cost of substantially lower success. Therefore, the constant selector provides a markedly better balance between target satisfaction and reference efficiency than any fixed global density.
The variable selector further reduces the average reference ratio to 16.2\%, but its success rate falls to 79.0\%. When equalised to the constant selector's reference budget, it reaches 92.0\% success. Thus, much of the gap between the selectors is attributable to reference count, while the constant selector retains a modest robustness advantage at equal storage.

\subsection{Constant selector vs.\ a fixed global baseline}

\Cref{tab:rar_absolute_with_refs} reports achieved RAR across all target Local Recall@1 and RAR combinations for each VPR method and dataset. The upper blocks show achieved RAR, while the lower blocks show the corresponding reference ratios. Our constant method selects a different subsampling factor for each operating condition, favouring the sparsest density whose estimated RAR satisfies the target. For our selector, most cells meet or exceed the target RAR, with the remaining shortfalls occurring only in settings where the maximum achievable RAR is itself below the target, as marked by an asterisk. The selected reference ratio increases as either the Local Recall@1 target or the target RAR becomes more demanding, reflecting the need for denser maps under stricter requirements.

By contrast, the fixed $k=4$ baseline produces several below-target cells, particularly for high Local Recall@1 and RAR requirements. For example, on Nordland with MixVPR and a Local Recall@1 target of 100\%, the fixed baseline achieves zero RAR across all target RAR values, whereas our selector reaches the maximum achievable RAR using the densest available map. The baseline is also strongly dataset-dependent. It performs well on Oxford RobotCar but fails more frequently on Nordland, particularly for MixVPR above 60\% Local Recall@1 and CosPlace under the strictest requirements.

\begin{figure*}[t]
\vspace*{0.2cm}
\centering
{\includegraphics[width=\linewidth, trim=1 18 1 14, clip]{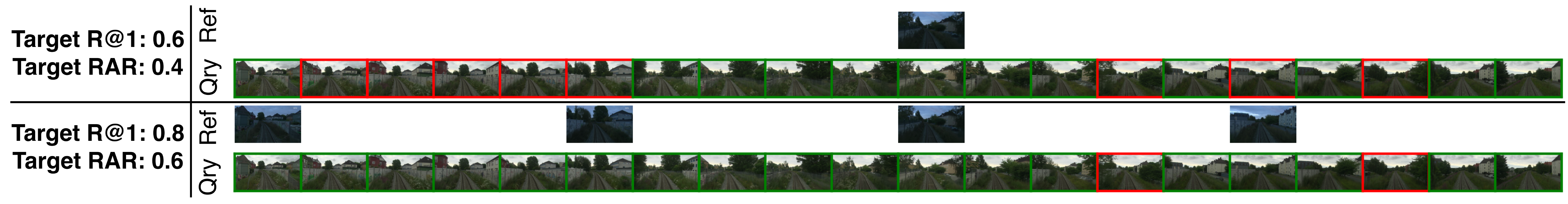}}
\caption{
Qualitative examples of the selected reference density under two user-specified operating points. Selected reference frames are shown above the query sequence; green and red borders indicate correct and incorrect top-1 matches, respectively.
[Top] For a target Local Recall@1 of 60\% and target RAR of 40\%, the method selects $k=15$, and the illustrated segment achieves 60\% Recall@1.
[Bottom] For a target Local Recall@1 of 80\% and target RAR of 60\%, the method selects $k=5$, and the illustrated segment achieves 90\% Recall@1.
}
\vspace*{-0.4cm}
\label{fig:qualitative}
\end{figure*}

\Cref{fig:qualitative} illustrates how the selected density adapts to the requested operating point. For the moderate target pair of 60\% Local Recall@1 and 40\% RAR, the method selects $k=15$, whereas the stricter target pair of 80\% Local Recall@1 and 60\% RAR requires $k=5$. These results show that reference-to-reference matching patterns are sufficiently informative to select densities that transfer effectively to unseen query data.

\subsection{Comparison with a reference-fusion method}
We also apply RAR to Bayesian Selective Fusion (BSF)~\cite{molloy2020intelligent}, a reference-refinement method with two reference traverses, autumn and spring, on Nordland (MixVPR). Fusing both full traverses, BSF stores 40,000 references at every operating point. At local Recall@1 and RAR of 80\%, it reaches a RAR of 100\%, versus our constant selector's achieved RAR of 88\% at $6\times$ less storage (6,667 references). At the strictest targets of RAR of 100\%, BSF achieved a RAR of 73\% (with 40,000 references) against our RAR of 63\% (with 20,000 references). Thus, fusion raises RAR only through added storage, whereas our selector provides an explicit target-driven storage-performance trade-off.

\subsection{Constant vs.\ variable reference sampling}
\label{res:constant_vs_variable}

\Cref{res:perf_overview} showed that the selectors approach similar success rates once their reference budgets are equalised. Here, we examine why the variable selector fails more often than the constant selector, and where the remaining shortfalls occur.

The variable selector assigns density independently to each segment using predictions obtained from Ref1 and Ref2, whereas final performance is measured on Ref1 and Qry1. Therefore, segment-level variation between Ref2 and Qry1 can change which local regions satisfy the target. The constant selector is less sensitive to individual segment-level prediction errors because its decision depends on the aggregate fraction of segments meeting the target under one global density.

The inflation margins in~\Cref{exp_setup:configs} partially compensate for this variation, allowing the unconstrained variable selector to use fewer references and often achieve an RAR closer to the specified threshold. Its failures occur primarily under the strictest requirements, particularly when Local Recall@1 or RAR equals 100\%. In these cases, few candidate densities are feasible and small Ref2-to-Qry1 differences can move individual segments below the target.
After budget equalisation, the variable selector's achieved RAR is close to that of the constant selector across most conditions, with residual shortfalls concentrated at the strictest RAR target of 100\%. This confirms that reference budget explains most, but not all, of the original performance gap.

\subsection{Per-segment reference density impact on VPR performance}

\begin{figure}
    \begin{subfigure}{\columnwidth}
        \centering
        \includegraphics[width=0.95\textwidth]{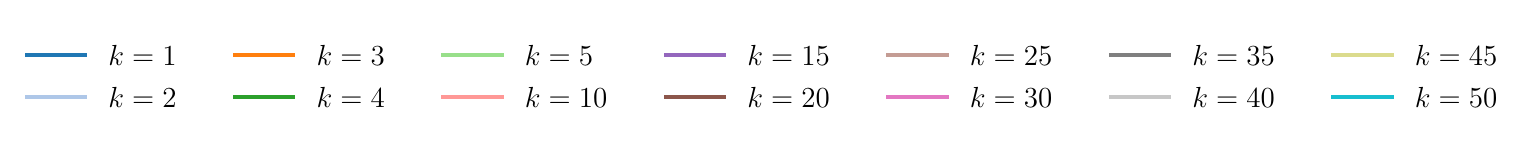}
    \end{subfigure}
    \vspace*{-0.1cm}
    \captionsetup[subfigure]{aboveskip=1.5pt,belowskip=0.5pt,labelformat=simple}
    \renewcommand*{\thesubfigure}{(\alph{subfigure})} %
    \centering
    \begin{subfigure}{0.8\columnwidth}
        \centering
        \includegraphics[width=\textwidth]{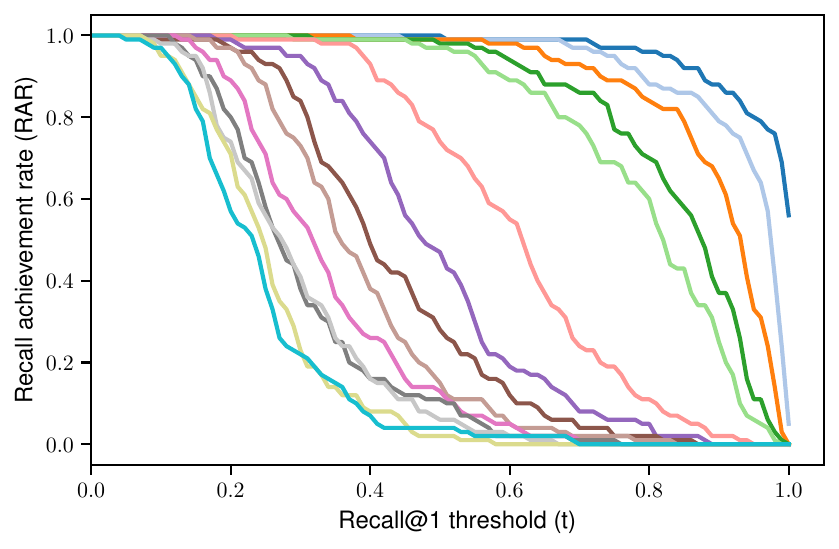}
    \end{subfigure}
    \vspace*{-0.3cm}
    \caption{
    Recall Achievement Rate (RAR) as a function of the target Local Recall@1 threshold for different fixed reference densities, from dense ($k=1$) to sparse ($k=50$), using CosPlace on Nordland. For each threshold, RAR is the proportion of segments whose Local Recall@1 meets or exceeds that threshold.
    }
    \label{fig:PerSeg_densitySR_Nordland}
    \vspace*{-0.3cm}
\end{figure}

We analyse how reference density affects VPR performance globally and locally, since this relationship underlies the density-selection problem.
As expected, denser reference sampling generally produces higher mean Recall@1, with performance decreasing as the map becomes sparser. The cumulative distributions in~\Cref{fig:PerSeg_densitySR_Nordland}, shown for CosPlace on Nordland, reveal substantial local variation. The fraction of segments meeting a given Local Recall@1 threshold changes markedly across subsampling factors, and the separation between curves confirms that reference density has a pronounced effect on segment-level performance.

Absolute Recall@1 also depends on the ground-truth tolerance. On Nordland, the relatively large tolerance of approximately $\pm 50$ frames allows even sparse maps to retain non-trivial performance. For MixVPR, $k=50$ achieves a mean Recall@1 of 25.8\%, compared with 98.3\% for $k=1$, while using only 2\% of the references, corresponding to 400 rather than 20,000 images. The effect of density nevertheless varies considerably across datasets, VPR methods, and individual segments.
Consequently, a single manually chosen density is unlikely to provide a reliable operating point across configurations. Our approach addresses this variation by choosing the sparsest sampling density predicted to meet the specified local requirements.

\subsection{Relationship between Recall@1 and RAR}

\begin{figure}
    \captionsetup[subfigure]{aboveskip=1.5pt,belowskip=0.5pt,labelformat=simple}
    \renewcommand*{\thesubfigure}{(\alph{subfigure})}
    \centering

    \vspace*{-0.2cm}
    \begin{subfigure}{0.8\columnwidth}
    \centering
    \includegraphics[width=\linewidth]{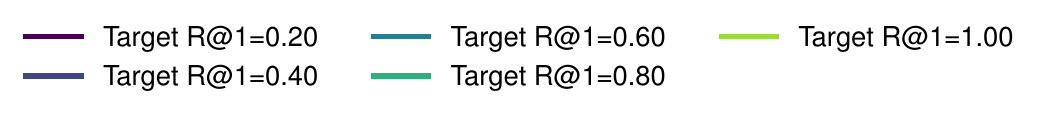}
    \end{subfigure}
    \vspace*{-0.1cm}
    \begin{subfigure}{0.98\columnwidth}
    \centering
    \includegraphics[width=\linewidth]{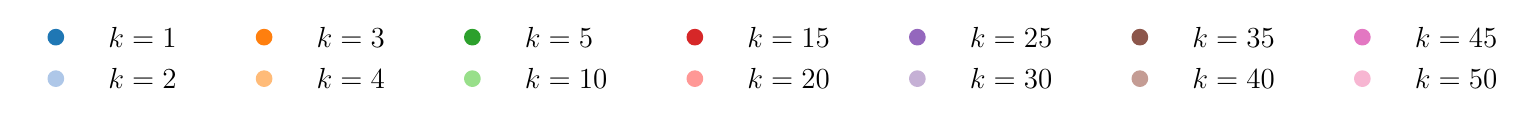}
    \end{subfigure}
    \begin{subfigure}{0.98\columnwidth}
        \centering
        \includegraphics[width=0.8\linewidth]{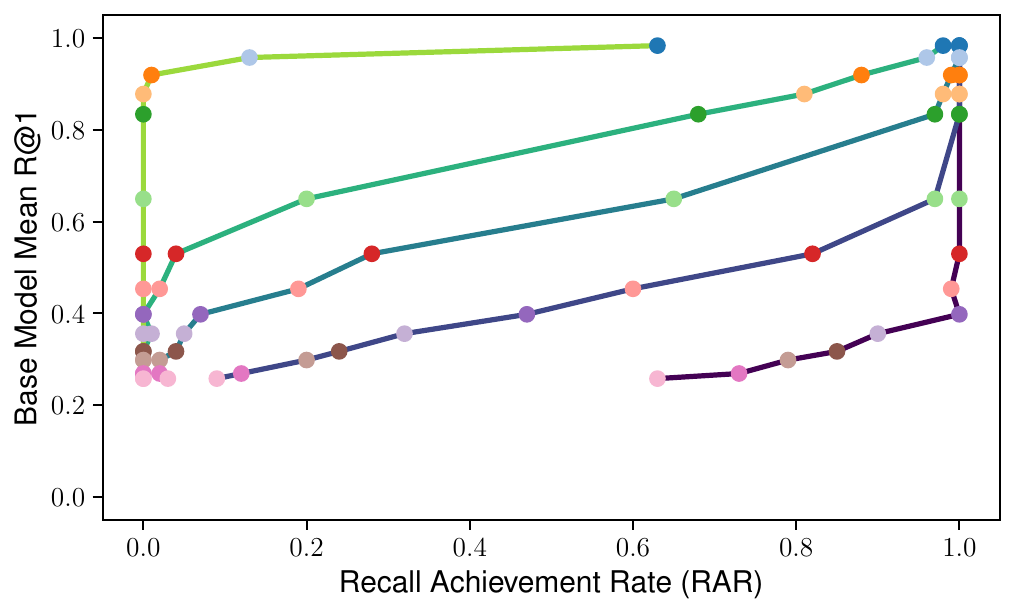}
    \end{subfigure}
    \vspace*{-0.3cm}
    \caption{
    Relationship between mean Recall@1 and Recall Achievement Rate (RAR) for fixed reference densities $k \in \mathcal{K}$ under Local Recall@1 targets of 20\%, 40\%, 60\%, 80\%, and 100\%, using MixVPR on Nordland. For a fixed density, mean Recall@1 remains unchanged, while RAR decreases as the Local Recall@1 target becomes stricter because fewer segments satisfy the requirement.
    }
    \label{fig:Rat1_vs_RAR}
    \vspace*{-0.3cm}
    
\end{figure}

\Cref{fig:Rat1_vs_RAR} demonstrates that \emph{high global Recall@1 does not imply high RAR for a given local target}. For example, at a per-segment target Recall@1 of 100\%, sampling with $k=3$ achieves a mean Recall@1 of 91.94\% but an RAR of only~1\%. Thus, despite near-perfect average performance, only 1\% of segments achieve 100\% Local Recall.
For a fixed VPR method, dataset, and subsampling factor, mean Recall@1 is unchanged by the selected Local Recall@1 target, whereas RAR decreases as that target becomes stricter. This causes the points for the same density to shift leftward in~\Cref{fig:Rat1_vs_RAR} as the target increases.

Therefore, the two metrics may appear correlated under permissive targets but diverge substantially under strict ones. RAR measures how broadly a required recall level is achieved across the environment, whereas Recall@1 measures only aggregate performance. Thus, similar global Recall@1 values can correspond to substantially different local reliability, motivating RAR as a complementary deployment metric.

\subsection{Ablation on swapping the reference traversals}
\label{res:abl_swapping}

To test sensitivity to the assignment of the two reference traversals, we swap their roles for MixVPR on Nordland. The original setting trains on Ref1 and Ref2 and evaluates on Ref1$^{\ast}$ and Qry1, whereas the swapped setting trains on Ref2 and Ref1 and evaluates on Ref2$^{\ast}$ and Qry1.
Both configurations achieve a 100\% success rate, with average reference ratios of 39.7\% and 45.2\% for the original and swapped orderings, respectively, indicating limited sensitivity to which traversal is assigned as Ref1.

\subsection{Ablation on segment length}
\label{res:abl_segment_len}

We vary the segment length from an average of 50 to 300 frames in increments of 50 using MixVPR on Nordland. For each setting, we report success rate and average reference ratio, defined as the retained reference count normalised by the densest map at $k=1$.
Success remains at or near 100\% across all segment lengths, showing low sensitivity to this choice. The average reference ratio generally decreases for longer segments because aggregating more frames smooths local variation and can make the RAR requirement achievable with sparser maps.
We use segments averaging 200 frames by default. 
It achieves 100\% success while retaining 39.7\% of the references. Although segments averaging 300 frames are marginally sparser at 39.6\%, very long segments can average over locally difficult regions and weaken the local interpretation that RAR is intended to provide.

\subsection{Recall estimation: predictor and lookup instantiations}
\label{res:pred_and_lookup}

Selecting a density ahead of deployment requires estimating, for each candidate factor $k$, the per-segment recall the reduced map will achieve. 
Our pipeline treats this as a modular stage. Any mechanism mapping reference-to-reference matching evidence to per-segment estimates $R_{i,k}$ can drive the same constant or variable selection policy. 
We instantiate two. The first is the Ridge predictor of~\Cref{method:pred}, the second is a direct lookup using the measured per-segment Ref1-to-Ref2 Recall@1 at each factor, with no learned model. 

As shown in~\Cref{fig:summary_fig_v2}, the two perform comparably, both reaching a $98.0\%$ success rate at average reference ratios of 23.4\% for the predictor and 24.8\% for direct lookup.
Thus, the informative signal lies in the reference-to-reference matching itself, and the outcome is robust to how that signal is converted into a density decision. 
The lookup is a strong, non-parametric option that is trivial to implement, while the predictor extends naturally to additional matching features or more complex reference-to-query relationships. Either can be adopted without changing the rest of the pipeline.

\subsection{Ablation study on predictor hyperparameter}
\label{res:abl_pred_hyp}

\begin{figure}[t]
\vspace*{0.2cm}
\centering{\includegraphics[width=0.85\linewidth]{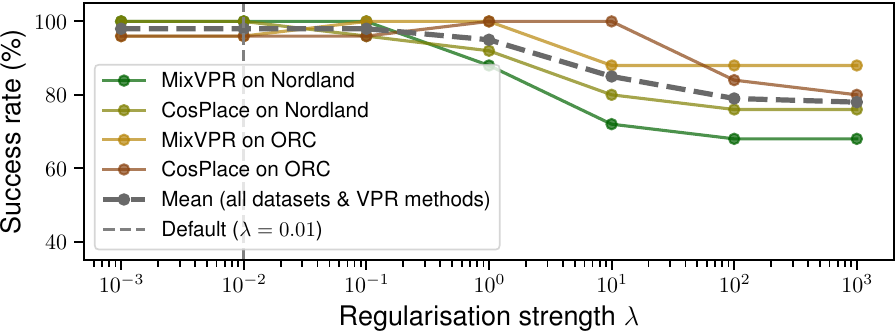}}
\vspace*{-0.2cm}
\caption{
Effect of Ridge-regression regularisation strength $\lambda$ on selector success rate across both VPR methods and datasets. Performance remains stable for $\lambda \in [10^{-3},1]$ and degrades under stronger regularisation. The default value $\lambda=0.01$, used in all experiments, lies within this stable region.
}
\vspace*{-0.4cm}
\label{fig:abl_pred_hyp}
\end{figure}

\Cref{fig:abl_pred_hyp} evaluates the effect of the Ridge regularisation strength $\lambda$. Increasing $\lambda$ strengthens coefficient shrinkage, while $\lambda=0$ corresponds to unregularised linear regression. Individual lines represent each VPR method and dataset combination, the dashed grey line shows their mean, and the vertical line marks the value $\lambda=0.01$ used in all experiments.
Performance remains stable across $\lambda \in [0,1]$, indicating low sensitivity within this range. Beyond $\lambda=1$, success decreases as stronger regularisation increasingly underfits the segment-level relationship.

\section{Discussion and Conclusions}

We have presented an approach that automatically selects a reference map density to meet user-specified Local Recall@1 and Recall Achievement Rate (RAR) targets, enabling performance-informed density selection before deployment rather than relying only on retrospective, globally averaged evaluation. Here, we discuss the assumptions and limitations of the approach, together with future research directions.

The approach assumes one-to-one spatial correspondence between the two reference traverses used for selection. This is well suited to repeated-route operations such as warehouse logistics, inspection, and last-mile delivery, where multiple traversals accumulate naturally during routine operation. Requiring a second reference traverse raises the practical bar relative to single-traversal pipelines, but the additional data enables local matching difficulty to inform density selection. Relaxing this requirement through synthetic augmentation or single-traversal uncertainty estimation is a promising future work. We note that one-to-one correspondence with the query traverse is required only for ground-truth evaluation.

Sparser subsampling reduces the spatial granularity of the available reference locations and can reduce localisation precision. In our experiments, the ground-truth tolerance remains sufficiently large relative to the candidate reference spacings to preserve valid matches under sparse sampling. This trades spatial precision for reduced storage and reliable place retrieval under the specified tolerance. Finer localisation could be recovered by interpolating query positions between neighbouring sparse reference frames.

Several extensions could broaden the framework. Although we target Local Recall@1, the same selection machinery could be applied to other task-relevant metrics, including mean, median, or worst-case translational and angular error. User requirements could vary spatially, allowing different targets in safety-critical regions, such as pedestrian crossings or decision points. 
More broadly, reference density is only one mechanism for meeting operational performance requirements. Others include varying the fidelity or number of sensor observations captured at each place. Coupled parameters such as segment size and map density could also be optimised jointly, while calibration could exploit more than two reference traverses. Overall, our results show that reference map density can be selected automatically from repeated traversals to satisfy local VPR requirements while avoiding unnecessary storage.

\bibliographystyle{IEEEtran}
\bibliography{references}

\end{document}